\documentclass[journal]{IEEEtran}

\usepackage{enumitem}
\usepackage{amsfonts}

\ifCLASSOPTIONcompsoc
  \usepackage[nocompress]{cite}
\else
  \usepackage{cite}
\fi

\ifCLASSOPTIONcompsoc
 \usepackage[caption=false,font=footnotesize,labelfont=sf,textfont=sf]{subfig}
\else
 \usepackage[caption=false,font=footnotesize]{subfig}
\fi

\ifCLASSINFOpdf
  \usepackage[pdftex]{graphicx}
  \graphicspath{{./figures/}}
  \DeclareGraphicsExtensions{.pdf,.jpeg,.png}
\else
\fi

\usepackage{booktabs}
\usepackage{makecell}
\usepackage{graphicx} 
\usepackage{pifont}
\usepackage{algorithm}
\usepackage{algorithmic}
\usepackage{amsmath}
\usepackage{amssymb}
\usepackage{wasysym}

\begin{document}
\title{Quantum Reinforcement Learning for Cost and
Delay Tradeoffs in Quantum Cloud Orchestration}

\author{An N. H. Phan,
Dang~Van~Huynh,~\IEEEmembership{Member,~IEEE,} Muhammad~Usman,~\IEEEmembership{Senior Member,~IEEE} and Hoa~T.~Nguyen,~\IEEEmembership{Member,~IEEE}

\IEEEcompsocitemizethanks{\IEEEcompsocthanksitem 
A. N. H. Phan and D. V. Huynh are with the Faculty of Computer Networks and Communications, University of Information Technology, Vietnam National University, Ho Chi Minh City, Quarter 34, Linh Xuan Ward, Ho Chi Minh City, Vietnam (e-mails: 23520029@gm.uit.edu.vn, danghv@uit.edu.vn).
}

\IEEEcompsocitemizethanks{\IEEEcompsocthanksitem 
M. Usman is with the Faculty of Information Technology, Monash University, Clayton, Victoria, Australia. The work was done while the author was at CSIRO (e-mail: muhammad.usman@monash.edu).  
} 

\IEEEcompsocitemizethanks{\IEEEcompsocthanksitem 
H. T. Nguyen is with the Quantum Systems Team, CSIRO, Clayton, Victoria, Australia (e-mail: hoa.nguyen@csiro.au).
}
}

\IEEEtitleabstractindextext{
\begin{abstract}
Quantum cloud computing, delivered through the quantum-as-a-service (QaaS) model, provides access to quantum computing resources. However, applying uniform time-based pricing across fundamentally heterogeneous quantum resources significantly complicates task orchestration, particularly when addressing the tradeoff between execution costs and system performance. While heuristic methods rely on predefined scheduling rules, classical deep reinforcement learning (DRL) models may require more trainable parameters in this setting. Motivated by the potential of parameterised quantum circuits (PQCs) as compact function approximators, we propose QRLQ, a cost-delay-aware quantum cloud scheduling framework integrating PQCs with a dueling double deep Q-network (D3QN) to dynamically account for both cost and delay. Our simulation results show that QRLQ achieves lower mean cost and delay than the heuristic baselines, achieving a 5-11\% lower mean cost relative to availability-based and rotation-based heuristics and reducing mean delay by 17\% and 82\% relative to the strongest and weakest heuristic baselines, respectively, while retaining execution fidelity within 2\% of a fidelity-greedy policy. Compared with the classical DRL baseline, QRLQ achieves comparable scheduling performance while using 72\% fewer trainable parameters. This work explores the feasibility of using QRL for task orchestration in quantum cloud environments and demonstrates its potential for cost-delay-aware quantum resource management.
\end{abstract}

}

\maketitle

\IEEEdisplaynontitleabstractindextext
\IEEEpeerreviewmaketitle

\ifCLASSOPTIONcompsoc
\IEEEraisesectionheading{\section{Introduction}\label{sec:introduction}}
\else
\section{Introduction}
\label{sec:introduction}
\fi

\IEEEPARstart{Q}{uantum} cloud computing is an emerging computing model that integrates quantum processing units (QPUs) into conventional cloud infrastructure, enabling on-demand access to quantum hardware through the quantum-as-a-service (QaaS) model \cite{nguyen2024quantum}. By reducing the need for individual organisations to acquire and maintain physical QPUs, QaaS lowers the barrier to quantum exploration and application development \cite{10643952}. Given its impact on accessibility and scalability, quantum cloud computing is expected to become one of the main ways to access quantum computing resources over the next decade \cite{9605297}.

Despite this potential, orchestrating QPU resources remains challenging due to hardware heterogeneity. Each QPU has different technical specifications, from high-level system metrics such as the number of qubits, quantum volume, and circuit layer operations per second \cite{wack2021qualityspeedscalekey} to low-level physical properties such as gate error rates and execution duration \cite{10.1145/3799898, 10.1145/3754598.3754641}. This heterogeneity directly affects execution fidelity, delay, and cost, making quantum task placement a multi-objective scheduling problem.

In the current noisy intermediate-scale quantum (NISQ) era \cite{preskill2018quantum}, maximising execution fidelity remains a primary objective for obtaining meaningful computational results \cite{10.1145/3799898}. However, under the commercial QaaS model \cite{bova2021commercial, NGUYEN2024281}, cost is also an important constraint. For example, advanced classical GPU instances typically cost tens of dollars per hour, whereas reserving dedicated access to the QuEra Aquila through Amazon Braket Direct costs \$2,500 per hour~\footnote{https://aws.amazon.com/braket/pricing (accessed: July 2026)}. the IBM Quantum Pay-as-you-go plan charges \$96 per QPU-minute~\footnote{https://www.ibm.com/quantum/products (accessed: July 2026)}. These prices show that quantum resource allocation should consider both cost and execution fidelity.

QaaS platforms use different pricing models that affect resource allocation decisions \cite{GHOSH2026108095}. Providers such as AWS Braket employ heterogeneous pricing based on the selected quantum device or request type, making cost differences among available quantum devices an important factor in resource allocation decisions. In contrast, platforms such as IBM Quantum operate under a homogeneous time-based pricing model, where different quantum resources are charged at the same rate and billed based on execution time. Because assigning requests to lower-performance hardware increases execution time and consequently cost, while concentrating workloads on high-performance hardware creates queues, scheduling under uniform pricing involves a direct cost-delay tradeoff. These queues increase delay \cite{giortamis2025qonductor}, which is also understood as the total completion time \cite{10643952}. Effective quantum cloud scheduling therefore needs to reduce execution time to lower cost while bounding delay.

\begin{figure}[!t]
\centering
\includegraphics[width=\linewidth]{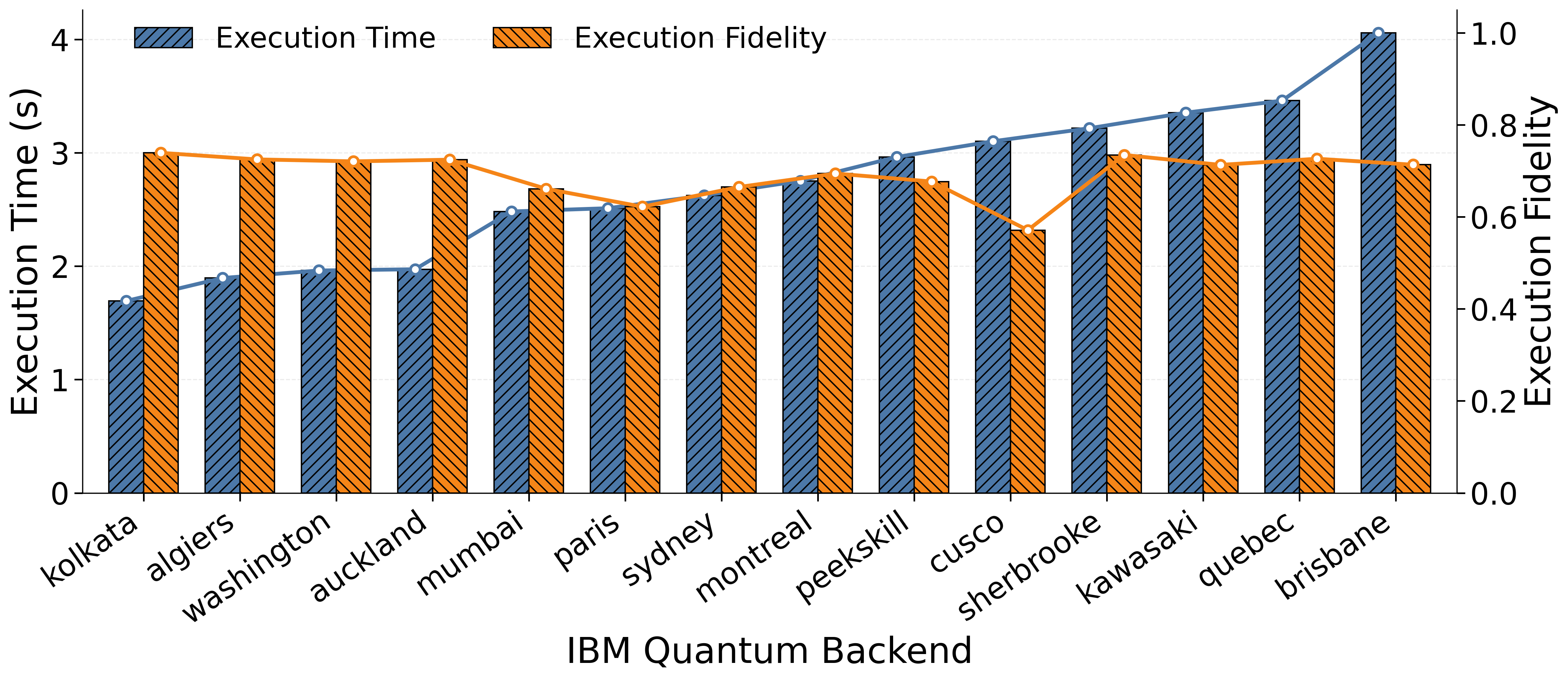}
\caption{Average execution time and execution fidelity across selected IBM Quantum simulated backends for the evaluated circuits with varying depths.}
\label{fig:execution_time_fidelity_tradeoff}
\end{figure}

Given that cost is directly proportional to QPU execution time, minimising execution time also minimises cost. Existing scheduling strategies often consider execution time as a secondary metric and prioritise execution fidelity as the primary objective \cite{10.1145/3799898, 10.1145/3754598.3754641}. This approach suggests a tradeoff between execution time and execution fidelity. However, across the evaluated IBM backends, shorter execution time did not coincide with lower execution fidelity. We evaluate three representative circuit families from the MQT Bench library \cite{quetschlich2023mqt}, namely GHZ, QAOA, and EfficientSU2, with varying circuit depths across selected IBM Quantum simulated backends. Fig.~\ref{fig:execution_time_fidelity_tradeoff} shows that the evaluated backends with shorter execution times generally exhibit relatively high execution fidelity, while those with longer execution times do not consistently exhibit higher fidelity. Therefore, selecting a higher-performance QPU may reduce both execution time and cost without reducing execution fidelity.

Existing work on quantum cloud resource orchestration relies mainly on heuristic strategies \cite{9605297}. Although some studies now consider cost as an optimisation objective \cite{GHOSH2026108095}, heuristic methods often lack the adaptability required to handle dynamic and stochastic environments. To overcome the limitations of static policies, recent work has adopted deep reinforcement learning (DRL) for adaptive scheduling in dynamic quantum cloud environments \cite{10643952, li2024moirai}. By optimising sequential scheduling decisions, DRL provides a suitable approach to address this problem. Building on advances in both DRL and quantum computing, quantum reinforcement learning (QRL) replaces the classical neural network of a DRL agent with parameterised quantum circuits (PQCs) \cite{skolik2022quantum}. By leveraging the potential parameter efficiency of PQCs, QRL provides a parameter-efficient function approximator for DRL and has demonstrated competitive performance in classical cloud resource allocation \cite{dai2025quantum}, suggesting its potential for the more constrained quantum cloud setting. 

To the best of our knowledge, no prior work applies QRL to quantum cloud resource orchestration under a uniform time-based pricing model. Moreover, current NISQ hardware is constrained by noise and gate errors, limiting the practical deployment of deep quantum circuits. Therefore, shallow PQCs are practically important for stable training on current quantum hardware \cite{prabhashana2025quantum, McClean2018}. Motivated by this research gap and the limitations of current NISQ hardware, we propose QRLQ, a \textbf{QRL}-based cost-delay-aware resource orchestration framework for \textbf{Q}uantum cloud environments with uniform time-based pricing. We also evaluate the proposed architecture under parameter perturbations in the context of current NISQ hardware, with promising results across the evaluated perturbation levels. This study explores the feasibility of applying QRL to quantum cloud resource orchestration, targeting future fault-tolerant and error-corrected environments, where the impact of operational noise is expected to be substantially reduced. The primary contributions of this study are as follows.

\begin{itemize}
    \item We formulate cost-delay-aware quantum task orchestration as an online sequential decision problem to jointly minimise execution and waiting times. To address the inherently stochastic dynamics of the environment, we model the orchestration process as a Markov Decision Process (MDP), thereby enabling reinforcement learning-based scheduling in dynamic quantum cloud settings.
    \item We propose QRLQ, a scheduling framework built on a quantum duelling double deep Q-network, for cost-delay-aware resource orchestration in quantum cloud environments. Our framework addresses the tradeoff between cost and delay by jointly minimising execution time and waiting time.
    \item We present the design of the hybrid quantum-classical network in our QRLQ framework, including the PQC configuration, quantum state encoding, and quantum-to-classical feature readout for duelling network-based scheduling across heterogeneous QPUs.
    \item We show that QRLQ outperforms heuristic strategies by reducing costs by 5--11\% and delay by 17--82\%, while retaining execution fidelity within 2\% of a fidelity-greedy policy. Compared with classical DRL, QRLQ lowers the number of model parameters by approximately 72\% while achieving comparable scheduling performance.
\end{itemize}

\textit{Paper structure and notations:} Section \ref{sec:related_work} reviews quantum resource orchestration approaches from static heuristics to DRL and discusses the potential of QRL for quantum cloud management. Section \ref{sec:system_model} formalises the system model and problem statement for heterogeneous quantum cloud orchestration with multiple objectives. Section \ref{sec:qrlq_framework} details the core methodology driving the hybrid QRLQ framework. Section \ref{sec:performance_evaluation} evaluates this framework alongside a discussion of simulation results. Section \ref{sec:conclusion} concludes the study with key insights and future research directions. Key mathematical notations are summarised in Table~\ref{tab:notation}.

\begin{table}[htbp]
\renewcommand{\arraystretch}{1.2}
\setlength{\tabcolsep}{2pt}
\centering
\caption{Summary of key notations.}
\label{tab:notation}
\begin{tabular*}{\columnwidth}{@{\extracolsep{\fill}}ll@{\hspace{8pt}}ll@{}}
    \toprule
    \textbf{Symbol} & \textbf{Description} &
    \textbf{Symbol} & \textbf{Description} \\
    \midrule
    $\mathcal{T}$ 
    & Set of QTasks.
    & $\mathcal{N}$
    & Set of QNodes. \\
    
    $\tau_i$
    & $i$-th QTask.
    & $n_j$
    & $j$-th QNode. \\
    
    $t_i^a$
    & Arrival time of $\tau_i$.
    & $t_j^r$
    & Next ready time of $n_j$. \\
    
    $T_j$
    & Operation-time mapping.
    & $E_j$
    & Operation-error mapping. \\
    
    $\mathcal{T}_{i,j}^{\mathrm{exec}}$
    & Execution time on $n_j$.
    & $\mathcal{T}_{i,j}^{\mathrm{wait}}$
    & Waiting time on $n_j$. \\
    
    $\mathcal{T}_{i,j}$
    & Total delay on $n_j$.
    & $\mathcal{F}_{i,j}$
    & Execution fidelity on $n_j$. \\
    
    $\mathcal{R}^{\mathrm{e}}$
    & Execution-time score.
    & $\mathcal{R}^{\mathrm{w}}$
    & Waiting-time score. \\
    
    $\Phi_{i,j}$
    & Overall scheduling score.
    & $C_i$ & Abstract circuit of $\tau_i$. 
    \\
    \bottomrule
\end{tabular*}
\end{table}

\section{Related Work}
\label{sec:related_work}
\subsection{Heuristic-based Orchestration Approaches}

\begin{figure*}[htbp]
\centering
\includegraphics[width=\linewidth,height=0.333\linewidth]{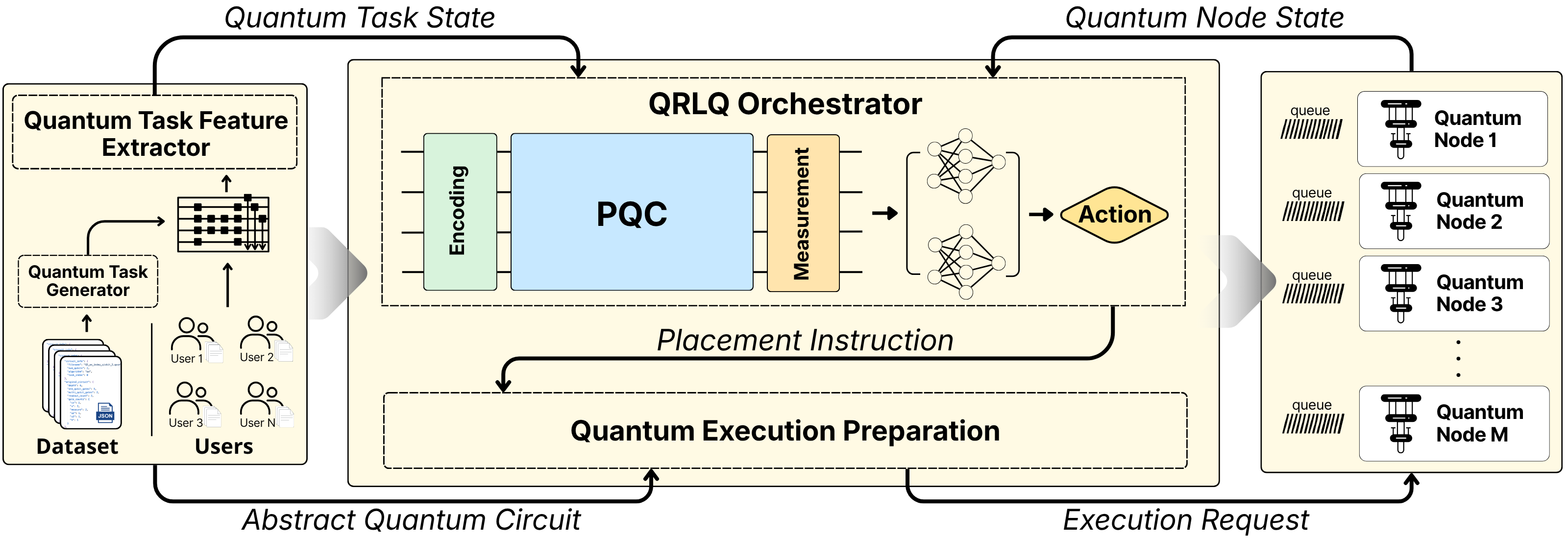}
\caption{System-level workflow of the proposed QRLQ framework for cost-delay-aware resource orchestration.}
\label{fig:system_architecture}
\end{figure*}

Early studies mainly used heuristics and mathematical optimisation. For example, adaptive scheduling based on historical IBM Quantum statistics aims to maximise execution fidelity under low-load conditions and minimise waiting time under high-load conditions \cite{9605297}. To improve execution fidelity, QuSplit uses a genetic algorithm to decompose quantum jobs based on backend noise levels \cite{li2025qusplit}.

Several studies in this category have considered cost. To handle demand uncertainty, two-stage stochastic programming methods have been applied to minimise the total cost, including reservation cost, usage cost, and penalties for exceeding the maximum waiting time \cite{kaewpuang2023stochastic}. Quantum-inspired particle swarm optimisation algorithms have also been proposed to minimise latency, energy consumption, and cost under execution-time pricing models \cite{GHOSH2026108095}. Similarly, multi-criteria decision-making methods have been integrated into systems such as Qonductor to provide resource plans that balance cost and execution fidelity \cite{giortamis2025qonductor}. However, rule-based, history-based, and mathematical programming approaches often struggle to adapt to dynamic quantum cloud environments, which are characterised by unpredictable workload arrivals and changing queue states.

\subsection{Learning-based Orchestration Approaches}
DRL has been used for quantum cloud orchestration to address the limitations of static heuristics, as it can learn scheduling policies from changing system states. DRLQ represents an early DRL-based framework in this direction, where Rainbow deep Q-networks (DQNs) are used to minimise total completion time and the number of rescheduling events \cite{10643952}. In the context of serverless quantum functions using the function-as-a-service model, graph convolutional networks combined with proximal policy optimisation (PPO) have been used to improve the distribution of quantum functions \cite{li2024moirai}. To address the challenge of executing large-scale circuits that exceed the capacity of a single QPU, PPO algorithms have also been applied to allocate sub-tasks across multiple QPUs \cite{10.1145/3754598.3754641}. This method maximises expected execution fidelity while learning its tradeoff with execution time and network communication latency. More recently, QFOR combines PPO with hardware calibration data to balance execution fidelity and execution time based on custom weights \cite{10.1145/3799898}. Although these DRL-based approaches are promising for quantum cloud orchestration, they mainly focus on maximising execution fidelity and minimising total completion time. These methods do not optimise cost during orchestration, even though cost is an important constraint for commercial QaaS platforms.

While DRL is a suitable approach for this orchestration problem, recent advances in quantum machine learning have made QRL a promising direction. Instead of relying on parameter-intensive dense neural networks, QRL employs PQCs as function approximators, providing a compact parameterisation with substantially fewer trainable parameters. Although QRL has been studied for multi-objective optimisation in other domains \cite{dai2025quantum, prabhashana2025quantum, wei2024quantum}, its specific application to quantum cloud resource orchestration has not yet been extensively investigated.

\section{System Model and Problem Formulation}
\label{sec:system_model}
\subsection{System Architecture}
Fig.~\ref{fig:system_architecture} illustrates the system-level workflow of the proposed QRLQ framework within a heterogeneous QaaS environment. The architecture consists of a centralised broker managed by the QRLQ orchestrator and a set of heterogeneous quantum nodes (QNodes) to process user-submitted quantum tasks (QTasks). Each QTask includes an abstract circuit with specific technical characteristics. Initially, a feature extractor processes this circuit to extract key task features. By evaluating these task features alongside the dynamic QNode states, the QRLQ core allocates each QTask to the QNode with the highest estimated action value. Then, the circuit is passed to the preparation stage before execution on the assigned QNode \cite{li2024moirai}.

In the NISQ era, the heterogeneity of quantum hardware results in different execution times for the same abstract circuit \cite{ravi2021quantum}. This variation can increase cost when a QTask is allocated to a QNode with a longer execution time. As a result, orchestrators may tend to assign more QTasks to a small number of high-performance QNodes. This concentration of tasks, together with the limited availability of high-performance quantum resources, can increase the waiting time of QTasks in queues before execution. In this study, we aim to jointly minimise execution time and waiting time using an equal-weight multi-objective formulation, thereby reducing both cost and delay.

\subsection{Quantum Task and Resource Models}
\subsubsection{Quantum Task Model}
In a quantum cloud environment, QTasks represent execution requests submitted by users to the system \cite{GHOSH2026108095}. Each QTask contains the technical parameters required for execution, such as the number of shots necessary for statistical sampling, and typically includes one or more gate-based quantum circuits. These circuits serve as low-level representations of a quantum algorithm \cite{9605297}. In this work, we model a QTask as a computational workload that requires adaptive resource orchestration to optimise execution performance. Furthermore, we assume each QTask consists of a single, independent abstract circuit.

As shown in the system architecture in Fig.~\ref{fig:system_architecture}, when a QTask arrives, the task feature extractor module analyses the abstract circuit to gather the essential parameters required for the scheduler's decision-making process. Formally, we consider a set of independent arriving QTasks, denoted as $\mathcal{T} = \{\tau_1, \tau_2, \dots, \tau_N\}$, where $N = |\mathcal{T}|$. Each QTask $\tau_i \in \mathcal{T}$ is formulated as

\begin{equation}
\label{eq:qtask_model} 
    \tau_i = (t_i^a, C_i, W_i, D_i, G_i^{(1q)}, G_i^{(2q)}, M_i, S_i),
\end{equation}

\noindent where $t_i^a$ denotes the arrival time of QTask $\tau_i$. The abstract circuit $C_i$ is represented as a directed acyclic graph (DAG). Before hardware-specific compilation, $C_i$ requires a circuit width $W_i$ and has a pre-compilation depth $D_i$. Furthermore, the gate-level execution cost is parameterised by the number of single-qubit and two-qubit gates, denoted as $G_i^{(1q)}$ and $G_i^{(2q)}$, respectively, along with the final measurement operations $M_i$. Finally, $S_i$ specifies the number of shots required to construct a statistically reliable probability distribution for the measurement outcomes.

\subsubsection{Quantum Resource Model}
In quantum cloud infrastructure, quantum computing resources are represented as QNodes, each abstracting a QPU together with its execution queue and hardware characteristics. Because of physical constraints in the NISQ era, these characteristics vary substantially across QNodes, including qubit capacity and qubit connectivity \cite{ravi2021quantum}. In modern hardware implementations, a physical quantum resource may include multiple QPUs and support multi-programming for parallel circuit execution \cite{9749894}. In this study, we focus on the single-QPU case and assume that each QNode abstracts a single dedicated QPU. Furthermore, a node provides exclusive access to one independent quantum circuit at any given time, provided that the required number of qubits does not exceed the available capacity of the target node. Execution is strictly non-preemptive \cite{10.1145/3799898, 10643952}.

Formally, we consider a heterogeneous set of QNodes available in the quantum cloud infrastructure, denoted as $\mathcal{N} = \{n_1, n_2, \dots, n_M\}$, where $M = |\mathcal{N}|$. Each QNode $n_j~\in~\mathcal{N}$ is modelled as the parameter tuple

\begin{equation}
\label{eq:qnode_model} 
    n_j = (t_j^{r}, Q_j, B_j, T_j, E_j),
\end{equation}

\noindent where $t_j^{r}$ denotes the expected next available time when the QNode $n_j$ completes its currently scheduled tasks and becomes \textbf{r}eady for new allocation. The parameter $Q_j$ represents the set of physical qubits available at the QNode, and $B_j$ specifies the native gate set supported by the hardware. Finally, hardware speed characteristics are described by the time mapping function $T_j: B_j \times \mathcal{A}_j \rightarrow \mathbb{R}_{>0}$, where $\mathcal{A}_j$ represents the set of all hardware-valid qubit operand configurations for executing supported operations on $n_j$. Similarly, the error function $E_j: B_j \times \mathcal{A}_j \rightarrow [0,1]$ represents the corresponding error probability of that execution.

\begin{figure}[htbp]
\centering
\includegraphics[width=\linewidth]{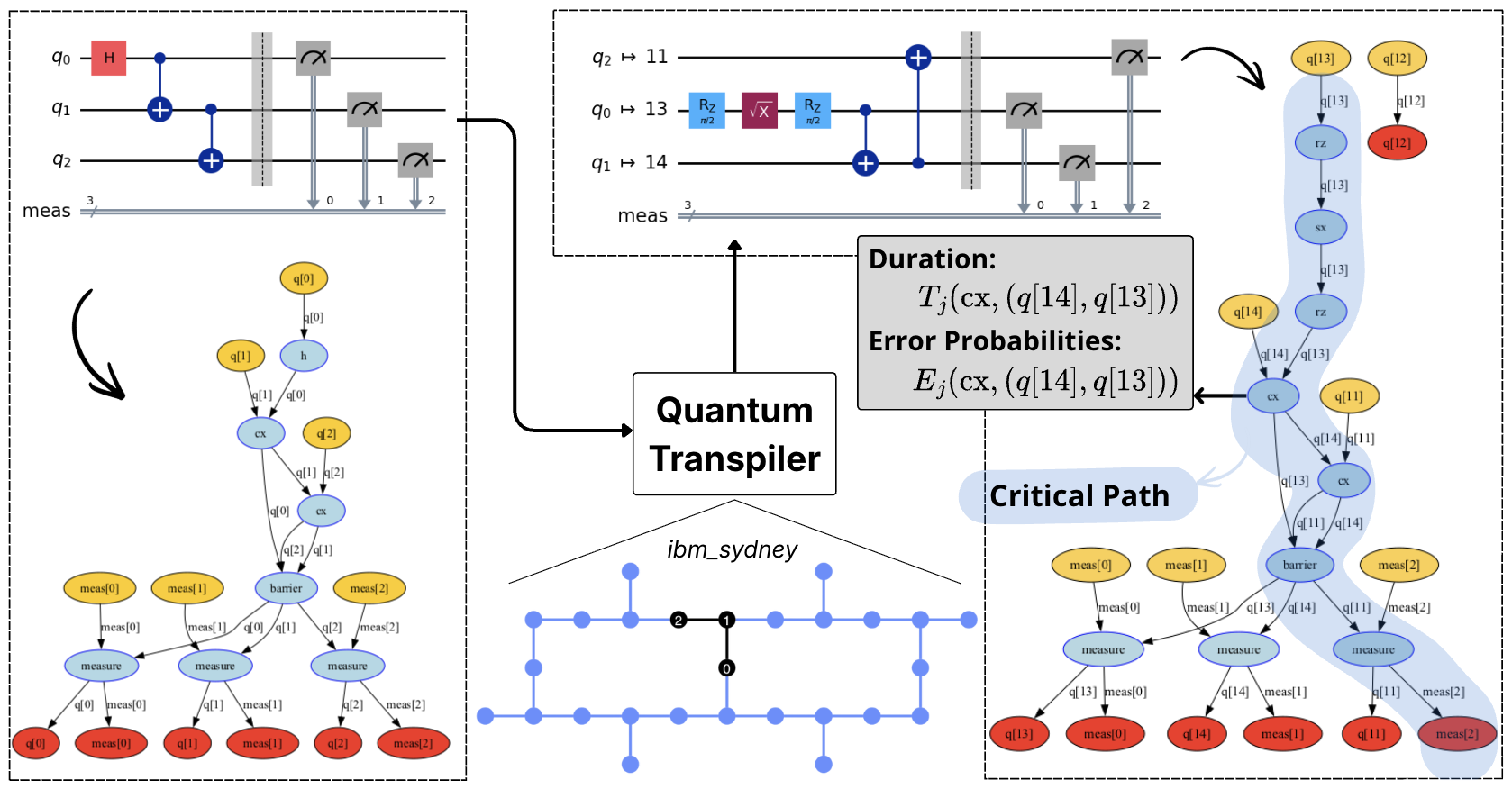}
\caption{Example of quantum circuit transpilation and its DAG representation for execution time and execution fidelity estimation.}
\label{fig:dag_circuit_transpilation}
\end{figure}

Figure \ref{fig:dag_circuit_transpilation} illustrates the processing of a QTask by quantum computing resources and the associated circuit compilation process. The abstract circuit, shown in the left panel, must be compiled and mapped to the physical qubit topology of the target quantum hardware, such as the \texttt{ibm\_sydney} system depicted in the centre panel. This procedure produces a transpiled circuit, represented as a DAG in the right panel. Notably, the DAG of the transpiled circuit enables execution time estimation via the critical path using operation times $T_j$, and supports fidelity evaluation through error probabilities $E_j$.

\subsection{Orchestration Performance Model}
As discussed in Section \ref{sec:introduction}, under the homogeneous pricing model, the cost is directly proportional to the actual execution time of the QPU. Let $\rho$ denote the unit price per unit time. For each accepted assignment, the cost of executing a QTask $\tau_i$ on a QNode $n_j$ is formally expressed as

\begin{equation}
\label{eq:cost_model} 
    \text{Cost}(\tau_i, n_j) = \rho \mathcal{T}_{i, j}^{\text{exec}},
\end{equation}

where the execution time $\mathcal{T}_{i, j}^{\text{exec}}$ is estimated directly by analysing the critical path of the transpiled circuit. We assume serial shot execution without reset or readout overheads, establishing a deterministic and consistent execution time model for training scheduling policies across heterogeneous backends, similar to \cite{10.1145/3799898}. Formally, the execution time can be calculated as

\begin{equation}
\label{eq:estimate_exectime} 
    \mathcal{T}_{i, j}^{\text{exec}} = S_i \sum_{g \in CP(C_{i,j}')} T_j(g, \vec{q}_g),
\end{equation}

\noindent where $C_{i,j}'$ is the transpiled circuit of $\tau_i$ mapped onto $n_j$, and $CP(C_{i,j}')$ denotes its critical path. $T_j(g, \vec{q}_g)$ represents the execution time of operation $g$ on the target qubit set $\vec{q}_g$ along this path, while $S_i$ is the number of shots requested by the user. In a quantum cloud environment, simple circuits with shallow depth and few qubits require significantly less execution time than complex circuits. This difference introduces the risk that scheduling strategies based mainly on execution time may favour simple circuits over complex circuits. To reduce this effect, we introduce a complexity bonus $\kappa_i \in (0, 1]$ for QTasks with higher circuit complexity \cite{10.1145/3799898}, which is formulated as

\begin{equation}
\label{eq:complexity_bonus} 
    \kappa_i = w_d \frac{D_i}{D_{\max}} + w_g \frac{O_i}{O_{\max}} + w_s \frac{S_i}{S_{\max}},
\end{equation}

\noindent where $D_i$ is the depth of the abstract circuit and $O_i$ is its total number of gates, calculated as $O_i = G_i^{(1q)} + G_i^{(2q)} + M_i$. $S_i$ represents the number of shots, while $w_d$, $w_g$, and $w_s$ act as adjustment coefficients where $w_d + w_g + w_s = 1$. The values $D_{\max}$, $O_{\max}$, and $S_{\max}$ denote the corresponding normalisation limits. Accordingly, the execution time performance score is determined as

\begin{equation}
\label{eq:exectime_score} 
    \mathcal{R}^{\text{e}} = (1-\eta) (1-\tilde{\mathcal{T}}_{i, j}^{\text{exec}}) + \eta \kappa_i,
\end{equation}

\noindent where $\tilde{\mathcal{T}}_{i, j}^{\text{exec}}$ denotes the execution time normalised by the maximum execution time observed in the dataset to ensure training stability, and $\eta$ is the complexity adjustment coefficient, set by default to $0.2$. Consequently, when a complex task receives a low performance score, $\kappa_i$ increases the total score $\mathcal{R}^{\text{e}}$ to reduce the disadvantage of high-complexity circuits, while execution time remains the main factor guiding the scheduler toward cost-efficient resource allocation.

The limited availability and high cost of quantum computing resources create competition for access to execution resources. We assume that each QNode maintains a queue where each QTask waits for execution. The waiting time $\mathcal{T}_{i, j}^{\text{wait}}$ of QTask $\tau_i$, calculated from its arrival at the system until the start of execution, is defined as

\begin{equation}
\label{eq:calcu_waittime} 
    \mathcal{T}_{i, j}^{\text{wait}} = \max(0, t_j^r - t_i^a),
\end{equation}

\noindent where $t_j^r$ is the time when QNode $n_j$ is ready for the next execution, and $t_i^a$ is the arrival time of QTask $\tau_i$. If $\mathcal{T}_{i, j}^{\text{wait}} = 0$, QTask $\tau_i$ can commence execution immediately. Subsequently, the next ready time $t_j^{r}$ of QNode $n_j$ is updated to $t_j^{r'} = \max(t_i^a, t_j^r) + \mathcal{T}_{i, j}^{\text{exec}}$. We define the overall delay experienced by QTask $\tau_i$ as its delay, formulated as

\begin{equation}
\label{eq:delay} 
    \mathcal{T}_{i, j} = \mathcal{T}_{i, j}^{\text{exec}} + \mathcal{T}_{i, j}^{\text{wait}}.
\end{equation} 

Finally, to evaluate the impact of delay on overall scheduling performance, let $\tilde{\mathcal{T}}_{i, j}^{\text{wait}}$ denote the waiting time normalised in the same manner as the execution time. The waiting time performance score is then defined as

\begin{equation}
\label{eq:waittime_score} 
    \mathcal{R}^{\text{w}} = 1 - \tilde{\mathcal{T}}_{i, j}^{\text{wait}}.
\end{equation}

We approximate the execution fidelity of QTask $\tau_i$ on QNode $n_j$ by accumulating the success probabilities of all physical operations in the transpiled circuit, following \cite{10.1145/3799898}. Formally, the execution fidelity is approximated as

\begin{equation}
\label{eq:estimate_execfidelity} 
    \mathcal{F}_{i,j} = \prod_{g \in C_{i,j}'} [1 - E_j(g, \vec{q}_g)],
\end{equation}

\noindent where $E_j(g, \vec{q}_g)$ is the physical error probability of operation $g \in C_{i,j}'$ on the target qubit set $\vec{q}_g$. Consequently, $[1 - E_j(g, \vec{q}_g)]$ represents the successful execution probability of that operation. 

\subsection{Problem Formulation}
We formally model quantum cloud task orchestration as an online sequential decision-making process. A sequence of QTasks $\mathcal{T}=\{\tau_1,\tau_2,\dots,\tau_N\}$ arrives at the system. Given a cluster of heterogeneous QNodes $\mathcal{N}=\{n_1,n_2,\dots,n_M\}$, the orchestration problem is to determine an orchestration policy $\pi: \mathcal{T} \rightarrow \mathcal{N}$ that assigns each QTask to a selected QNode. For each task assignment, the orchestration performance score is defined as

\begin{equation}
\label{eq:overall_score}
    \Phi_{i,\pi(\tau_i)}
    =
    \beta
    \mathcal{R}^{\text{e}}_{i,\pi(\tau_i)}
    +
    (1-\beta)
    \mathcal{R}^{\text{w}}_{i,\pi(\tau_i)},
\end{equation}

\noindent where $\beta \in [0,1]$ is the strategy adjustment parameter controlling the tradeoff between execution time and waiting time. In this study, $\beta$ is fixed to $0.5$ to assign equal priority to both optimisation objectives.

The online orchestration objective is to maximise the cumulative scheduling performance, which can be formulated as

\begin{subequations}
    \label{eq:orchestration_problem}
    \begin{align}
        \max_{\pi}\quad
        &\sum_{i=1}^{N}
        \Phi_{i,\pi(\tau_i)},
        \label{eq:orchestration_objective}\\
        \text{s.t.}\quad
        &
        W_i
        \le
        |Q_{\pi(\tau_i)}|,
        \qquad
        \forall \tau_i\in\mathcal{T},
        \label{eq:qubit_capacity_constraint}\\
        &
        \begin{aligned}
            (g, \vec{q}_g) &\in B_{\pi(\tau_i)} \times \mathcal{A}_{\pi(\tau_i)}, \forall (g, \vec{q}_g) \in C_{i,\pi(\tau_i)}'.
        \end{aligned}
        \label{eq:gate_compatibility_constraint}
    \end{align}
\end{subequations}

Constraint~\eqref{eq:qubit_capacity_constraint} ensures that the selected QNode provides sufficient physical qubits for the assigned QTask. Constraint~\eqref{eq:gate_compatibility_constraint} ensures that the transpiled circuit is compatible with the native operations and valid qubit operands of the assigned QNode.

\subsection{Markov Decision Process (MDP) Formulation}
\label{sec:mdp}

\begin{figure*}[!t]
\centering
\includegraphics[width=\linewidth]{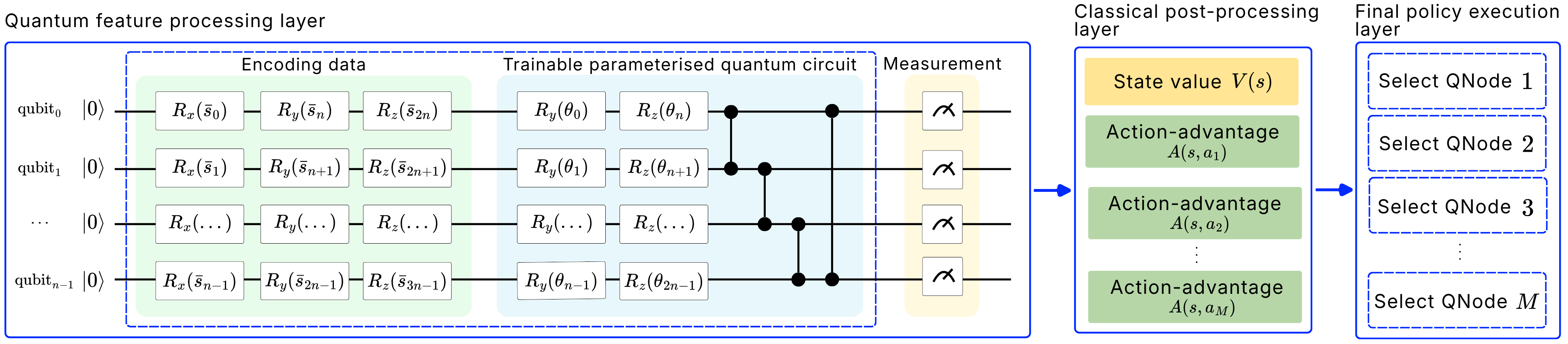}
\caption{The proposed cost-delay-aware QRLQ framework for quantum cloud resource orchestration, with eight qubits used in our simulations.}
\label{fig:proposed_qrlq_architecture}
\end{figure*}

The QRLQ framework models the resource scheduling process as an MDP \cite{10.1145/3799898, dai2025quantum}. This allows the system to use reinforcement learning methods to approximate a high-performing action-value function instead of using a fixed scheduling rule. This MDP is represented by the tuple $(\mathbf{S}, \mathbf{A}, \mathbf{P}, \mathbf{R}, \gamma)$, with $\mathbf{S}$ denoting the continuous state space and $\mathbf{A}$ representing the discrete action space corresponding to hardware allocation decisions. The transition function $\mathbf{P}$ defines the state transition of the environment, including the evolution from state $s_t$ to $s_{t+1}$ by moving to the subsequent QTask $\tau_{i+1}$ in the submission sequence and updating the available time $t_j^{r'}$ of the selected QNode. Finally, $\mathbf{R}$ is the reward function guiding the optimisation, and $\gamma \in [0, 1)$ is the discount factor that controls short-term and long-term returns throughout the scheduling decision process.

At each decision step $t$, triggered by the arrival of QTask $\tau_i$, the QRLQ agent observes the state $s_t \in \mathbf{S}$, selects an action $a_t$ according to the policy $\pi(a_t | s_t)$, receives an immediate scalar reward $r_t \in \mathbb{R}$, and transitions to the next state $s_{t+1}$.

\textbf{State Space $\mathbf{S}$:} At decision step $t \in \mathbb{T}$, the observed state $s_t \in \mathbf{S}$ is constructed by concatenating two distinct feature vectors encoding the characteristics of the incoming QTask $\tau_i$ and the hardware configuration of the available QNodes. Mathematically, the complete state vector $s_t$ is expressed as

\begin{equation}
\label{eq:state_vector} 
    s_t = (\mathbf{f}^{\tau_i}_t, \mathbf{f}^{\mathcal{N}}_t).
\end{equation}

Specifically, task features are represented by the vector $\mathbf{f}^{\tau_i}_t \in \mathbb{R}^p$, where $p$ denotes the number of extracted features. These include the composite complexity, computed as the product of the number of shots and circuit layers, along with the quantum load, which encompasses the number of single-qubit gates, two-qubit gates, and measurement operations, as introduced in Eq.~\eqref{eq:qtask_model}. Meanwhile, the feature matrix $\mathbf{f}^{\mathcal{N}}_t \in \mathbb{R}^{M \times o}$ comprises the feature vectors of the $M$ available QNodes, where each node is characterised by $o$ features. These hardware features include the relative ready time of the QNode and performance parameters such as the average execution duration of gates and corresponding measurement operations, derived from Eq.~\eqref{eq:qnode_model}. All features are scaled to $[0,1]$ and then mapped to rotation angles in $[0,\pi]$ before being encoded into the parameterised quantum circuit. Finally, the overall state space is defined as

\begin{equation}
\label{eq:state_space} 
    \mathbf{S} = \{s_t \mid s_t = (\mathbf{f}^{\tau_i}_t, \mathbf{f}^{\mathcal{N}}_t), \forall t \in \mathbb{T}\}.
\end{equation}

\textbf{Action Space $\mathbf{A}$:} This discrete space defines the choice of a candidate QNode for allocating the incoming QTask based on the learned policy. At each decision step $t$, the agent selects an action $a_t \in \{0, 1, \dots, M - 1\}$, where $M = |\mathcal{N}|$ denotes the total number of available QNodes. Specifically, $a_t = j$ represents assigning the task to the $j$-th QNode $n_j \in \mathcal{N}$. Consequently, the action space is mapped directly to the set of QNodes, expressed as $\mathbf{A} = \mathcal{N}$.

\textbf{Reward Function $\mathbf{R}$:} The immediate reward $r_t$ is formulated to directly align with the optimisation objective of the scheduling strategy. Upon allocating the QTask $\tau_i$ to the $j$-th QNode at decision step $t$, this reward integrates the performance scores of both execution time and waiting time, governed by a tradeoff weighting parameter $\beta \in [0, 1]$. Formally, the calculation is expressed as

\begin{equation}
\label{eq:reward} 
    r_t = \beta \mathcal{R}^{\text{e}} + (1-\beta) \mathcal{R}^{\text{w}}.
\end{equation}

\section{The Proposed QRLQ Framework}
\label{sec:qrlq_framework}

\subsection{Overall QRLQ Architecture}
The proposed QRLQ framework introduces a hybrid architecture that integrates Double DQN and Duelling DQN and replaces the dense feature extractor with a PQC to address cost-delay-aware resource scheduling. Double DQN separates action selection from action evaluation using two neural networks with the same structure, which reduces the overestimation bias in DQN and improves training stability \cite{van2016deep}. The Duelling DQN architecture \cite{wang2016dueling} separates the state-value function $V(s)$ from the advantage function $A(s, a)$. This separation allows the model to estimate both the general value of a scheduling state and the relative importance of each allocation action, thereby facilitating more effective action selection in environments with heterogeneous QTask characteristics. The framework replaces the dense feature extractor of a classical D3QN with a parameterised quantum circuit \cite{jerbi2021parametrized, skolik2022quantum}. As shown in Section \ref{sec:computational_complexity}, this reduces the trainable parameter count from 758 to 214 while maintaining classical scheduling performance.

Fig.~\ref{fig:proposed_qrlq_architecture} illustrates the high-level architecture of the proposed QRLQ framework. First, the environment state space is transformed into a $d$-dimensional vector with elements normalised to map onto the rotation angles of each qubit within an $n$-qubit system. Then, a trainable PQC processes the encoded state and applies entangling operations across adjacent qubits to transform the encoded feature vectors before measurement. To support scalable circuit configurations, both the data encoding block and the PQC are formulated with a configurable number of layers $l$, allowing the architecture to adopt a data re-uploading scheme \cite{perez2020data} when $l>1$. Quantum measurements are then performed to map the quantum state back to classical expectation values. The output from these quantum measurements serves as the input feature vector for the classical post-processing layer. Within this layer, the data flow is separated into two independent computational streams. One stream estimates the state value, whereas the other evaluates the advantage for each specific scheduling action. The outputs from both streams are then combined to compute the final action-value function $Q(s, a)$ for each candidate action. Based on these estimated values, the policy selects the action with the highest estimated value to allocate the QTask to the selected QNode $n_j$ from the $M$ available options.

\subsection{Parameterised Quantum Circuit (PQC) Design}
Algorithm \ref{alg:qrlq_impl} presents the detailed implementation of this hybrid architecture to compute the action-value function. To align with the information processing characteristics of the PQC, the features $s_i$ from the environment state $\mathbf{S}$ are normalised into $\bar{s}_i$ to correspond with rotation angles within the interval $[0, \pi]$, thereby mapping the information onto a hemisphere of the Bloch sphere \cite{costello2021reconstruction}. The normalised data is embedded into the quantum circuit via rotation gates $R_x$, $R_y$, and $R_z$. Following the encoding layer, the variational layer applies parameterised single-qubit rotations along the $y$ and $z$ axes to every qubit \cite{skolik2022quantum}. The corresponding unitary operator is defined as

\begin{equation}
\label{eq:variational_layer}
\begin{aligned}
    U_{\mathrm{var}}(\boldsymbol{\theta})
    =
    \bigotimes_{i=0}^{n-1}
    R_z^{(i)}(\theta_{i,z})
    R_y^{(i)}(\theta_{i,y}),
\end{aligned}
\end{equation}

\noindent where the rotation angles $\theta_{i,y}$ and $\theta_{i,z}$ are trainable parameters that determine the local unitary transformations applied to the $i$-th qubit. Consequently, the overall quantum state of the circuit after the variational layer is expressed as

\begin{equation}
\label{eq:quantum_state_variational}
\begin{aligned}
    |\psi''\rangle
    =
    U_{\mathrm{var}}(\boldsymbol{\theta})
    |\psi'\rangle,
\end{aligned}
\end{equation}

\noindent where $|\psi'\rangle$ and $|\psi''\rangle$ denote the global quantum states of the register after the encoding layer and after passing through the variational layer, respectively.

Although these variational operations provide value function approximation capabilities, they act locally on individual qubits. To enable the model to capture complex correlations among the features of quantum cloud resources \cite{dai2025quantum}, quantum entanglement is implemented via CZ gates that connect adjacent qubits in a ring topology. The output quantum state $|\Psi_{out}\rangle$ is formulated as

\begin{equation}
\label{eq:quantum_output_state}
\begin{aligned}
    |\Psi_{out}\rangle
    =
    \left(
    \prod_{i=0}^{n-1}
    CZ^{(i,(i+1)\bmod n)}
    \right)
    |\psi''\rangle,
\end{aligned}
\end{equation}

\noindent where the operator $CZ^{(i,(i+1)\bmod n)}$ entangles the $i$-th control qubit with its adjacent target qubit, forming a ring topology that captures correlations among neighbouring resource features. 

\begin{algorithm}[!t]
    \caption{Proposed QRLQ implementation.}
    \label{alg:qrlq_impl}
    \renewcommand{\algorithmicrequire}{\textbf{Input}}
    \renewcommand{\algorithmicensure}{\textbf{Output}}
    \begin{algorithmic}[1]
        \REQUIRE State $s$, number of qubits $n$, layers $l$, complete set of trainable parameters $\theta = \{\theta_{i, y}, \theta_{i, z}, \theta_V, \theta_A\}$
        
        \STATE Normalise input features $\bar{s}_i$ for all elements in $s$
        \STATE Initialise quantum state $|\psi\rangle \leftarrow |0\rangle^{\otimes n}$
        
        \FOR{$k = 1, \dots, l$}
            
            \STATE \textbf{\# data encoding}
            \FOR{each qubit $i = 0$ to $n-1$}
                \STATE Apply $R_z(\bar{s}_{2n+i}) R_y(\bar{s}_{n+i}) R_x(\bar{s}_i)$ to qubit $i$
            \ENDFOR
            
            \STATE \textbf{\# parameterised quantum circuit}
            \FOR{each qubit $i = 0$ to $n-1$}
                \STATE Apply $R_y(\theta^{k}_{i,y})$ and $R_z(\theta^{k}_{i,z})$
            \ENDFOR
            
            \FOR{each qubit $i = 0$ to $n-2$}
                \STATE Apply entanglement $CZ(i, i+1)$
            \ENDFOR
            \STATE Apply $CZ(n-1, 0)$
            
        \ENDFOR
        \STATE Obtain final quantum state $|\Psi_{out}\rangle$
        
        \STATE \textbf{\# measurement}
        \STATE $O \leftarrow [\langle Z_0 \rangle, \dots, \langle Z_{n-1} \rangle]$
        
        \STATE \textbf{\# duelling network}
        \STATE $V \leftarrow f_V(O, \theta_V)$
        \STATE $A \leftarrow f_A(O, \theta_A)$
        \STATE Compute $Q(s,a, \theta)$ using Eq.~\eqref{eq:action_value}
        
        \ENSURE $Q(s,a, \theta)$ for all actions
    \end{algorithmic}
\end{algorithm}

\subsection{Value Observation and Duelling Output}
To extract classical information from the quantum state transformed by the PQC layers, an expectation value measurement using the Pauli-Z operator is applied across all $n$ qubits. The expectation value on the $i$-th qubit is defined as

\begin{equation}
\label{eq:single_qubit_measurement}
\begin{aligned}
    \langle O_i \rangle = \langle \Psi_{out} | Z_i | \Psi_{out} \rangle,
\end{aligned}
\end{equation}

\noindent where $Z_i$ represents the Pauli-Z operator acting on the $i$-th qubit. This process generates a feature vector expressed as

\begin{equation}
\label{eq:measurement_vector}
\begin{aligned}
    O = [\langle O_0 \rangle, \langle O_1 \rangle, \dots, \langle O_{n-1} \rangle]^T \in [-1, 1]^n.
\end{aligned}
\end{equation}

Consequently, the vector $O$ is processed through a classical post-processing layer. Specifically, the state-value stream $V(O; \theta_V)$ parameterised by $\theta_V$ evaluates the overall value of the current state, whereas the advantage stream $A(O, a; \theta_A)$ parameterised by $\theta_A$ estimates the relative advantage of each action. The action-value function of the hybrid architecture is formally defined as

\begin{equation}
\label{eq:action_value}
\begin{aligned}
Q(s, a; \theta) = V(O; \theta_V)
+ \Big(& A(O, a; \theta_A) \\
       & - \frac{1}{|\mathcal{A}|} \sum_{a'} A(O, a'; \theta_A) \Big),
\end{aligned}
\end{equation}

\noindent where $a'$ represents a feasible action and $|\mathcal{A}| = |\mathcal{N}|$ denotes the action space size equal to the number of available QNodes within the system. The notation $\theta$ denotes the complete set of trainable parameters for the hybrid model, including the classical weights $\theta_V$ and $\theta_A$ alongside the quantum rotation angles $\theta_{i, y}$ and $\theta_{i, z}$ within the PQC.

\begin{table*}[!t]
\renewcommand{\arraystretch}{1.4}
\setlength{\tabcolsep}{10pt}
\centering
\caption{Simulation hyperparameter settings across policies \cite{skolik2022quantum, dai2025quantum}.}
\label{tab:hyperparameter_settings}
\begin{tabular}{lccc}
    \toprule
    \textbf{Settings} & \textbf{QRLQ} & \textbf{QDQN} & \textbf{D3QN} \\
    \midrule
    Architecture &
        \makecell[c]{1 re-upload layer \\ 16 variational gates \\ Duelling: 8 neurons} &
        \makecell[c]{10 re-upload layers \\ 160 variational gates} &
        \makecell[c]{1 hidden layer (LayerNorm) \\ 16 neurons \\ Duelling: 8 neurons} \\
    Learning rate &
        \makecell[c]{Angles: 0.001} &
        \makecell[c]{Angles: 0.001 \\ Scaling: 0.001} &
        0.001 \\
    \textbf{Training parameters} &
        \textbf{214} & \textbf{170} & \textbf{758} \\
    Shared training settings: &
        \multicolumn{3}{c}{%
        \makecell[c]{%
        Update: Every 10 steps, \quad
        $\gamma=0.99$, \quad
        Batch size: 64, \quad
        Exploration: 65\% episodes, \quad
        $\tau=0.001$.}} \\
    \bottomrule
\end{tabular}
\end{table*}

\begin{algorithm}[!t]
    \caption{Proposed QRLQ training workflow.}
    \label{alg:qrlq_train_workflow}
    \renewcommand{\algorithmicrequire}{\textbf{Input}}
    \begin{algorithmic}[1]
        \REQUIRE Replay buffer capacity $C$, minibatch size $B$, learning rate $\alpha$, discount factor $\gamma$, soft update coefficient $\tau$, update freq $f$, total episodes $E$.
        
        \STATE Initialise replay buffer $\mathcal{D} \leftarrow \emptyset$, global time step $t \leftarrow 0$
        \STATE Initialise $\theta_{onl}$ and $\theta_{tar}$, set $\theta_{tar} \leftarrow \theta_{onl}$
        
        \FOR{each episode $e = 1, \dots, E$}
            \STATE Decay exploration rate $\epsilon$
            \STATE $s \leftarrow \text{env.reset()}$
            
            \FOR{each time step}
                \STATE Compute $Q(s, a, \theta_{onl})$ using Algorithm~\ref{alg:qrlq_impl}
                
                \STATE Select action $a \sim \epsilon\text{-greedy}(Q)$
                \STATE Execute $(r, s', d) \leftarrow \text{step}(a)$
                \STATE Store transition $\mathcal{D} \leftarrow \mathcal{D} \cup \{(s, a, r, s', d)\}$, $t \leftarrow t + 1$
                
                \IF{$t \bmod f = 0$ \AND $|\mathcal{D}| \ge B$}
                    \STATE Sample minibatch $(s_i, a_i, r_i, s'_i, d_i)$
                    \STATE $a^* \leftarrow \arg\max_{a'} Q(s', a', \theta_{onl})$
                    \STATE Compute target $Y$ using Eq.~\eqref{eq:td_target}
                    
                    \STATE Optimise $\theta_{onl}$ via Eq.~\eqref{eq:loss}
                    \STATE Apply gradient clipping $\|\nabla\| \leq 1$
                    \STATE Soft update $\theta_{tar} \leftarrow \tau \theta_{onl} + (1-\tau)\theta_{tar}$
                \ENDIF
                
                \STATE $s \leftarrow s'$
                
                \IF{$d$ is True}
                    \STATE \textbf{break}
                \ENDIF
            \ENDFOR
        \ENDFOR
    \end{algorithmic}
\end{algorithm}

\subsection{Learning Stabilisation and Training Strategy}
The main objective of the QRLQ framework is to approximate the optimal action-value function to maximise the long-term cumulative reward and support resource scheduling decisions under dynamic resource conditions. Following the MDP formulation, the interaction process at each decision step $t$ generates a transition tuple $(s_t, a_t, r_t, s_{t+1})$, which is stored in an experience replay memory. The training procedure is summarised in Algorithm \ref{alg:qrlq_train_workflow}. To optimise the model, the system randomly samples mini-batches from memory to break temporal correlation and stabilise training. Based on these data samples, the QRLQ framework extends the double Q-learning architecture \cite{van2016deep} to reduce the overestimation bias in the DQN algorithm. The framework uses two structurally identical neural networks comprising an online network parameterised by $\theta_{onl}$ to select the optimal action in the subsequent state, and a target network parameterised by $\theta_{tar}$ to evaluate the value of that action. The temporal difference (TD) target $Y_t$ is formulated as

\begin{equation}
\label{eq:td_target}
\begin{aligned}
    a^* &= \arg\max_{a'} Q(s_{t+1}, a'; \theta_{onl}), \\
    Y_t &= r_t + \gamma (1 - d_t) Q(s_{t+1}, a^*; \theta_{tar}),
\end{aligned}
\end{equation}

\noindent where $d_{t} \in \{0, 1\}$ denotes the terminal flag, which equals 1 if $s_{t+1}$ is the final state of the scheduling sequence and 0 otherwise, thereby disabling the future value term.

For a batch of ${B}$ transitions sampled randomly from the experience replay memory, let $\delta_i = Q(s_i, a_i; \theta_{onl})~-~Y_i$ denote the TD error of the $i$-th sample. The loss function $L(\theta_{onl})$ \cite{mnih2015human} is given by

\begin{equation}
\label{eq:loss}
\begin{aligned}
    L(\theta_{onl}) = \frac{1}{B} \sum_{i=1}^{B} \begin{cases} \frac{1}{2} \delta_i^2 & \text{if } |\delta_i| < 1 \\ |\delta_i| - \frac{1}{2} & \text{if } |\delta_i| \ge 1. \end{cases}
\end{aligned}
\end{equation}

To ensure stability during the optimisation process, gradient clipping is applied such that $||\nabla_{\theta_{onl}} L|| \le 1.0$. After gradient clipping, all classical and quantum parameters within $\theta_{onl}$ are updated using a gradient-based optimisation method with a learning rate $\alpha$, which is expressed as

\begin{equation}
\label{eq:online_update}
\begin{aligned}
    \theta_{onl} \leftarrow \theta_{onl} - \alpha \nabla_{\theta_{onl}} L.
\end{aligned}
\end{equation}

Finally, a soft-target update strategy is implemented to incrementally update the target network's weights. This update rule is formulated as

\begin{equation}
\label{eq:target_update}
\begin{aligned}
    \theta_{tar} \leftarrow \tau \theta_{onl} + (1 - \tau) \theta_{tar},
\end{aligned}
\end{equation}

\noindent where $\tau \ll 1$ denotes the soft update coefficient. This update reduces unstable Q-value estimates and helps stabilise the training process of the QRLQ framework.

\begin{figure}[htbp]
\centering
\includegraphics[width=0.96\linewidth]{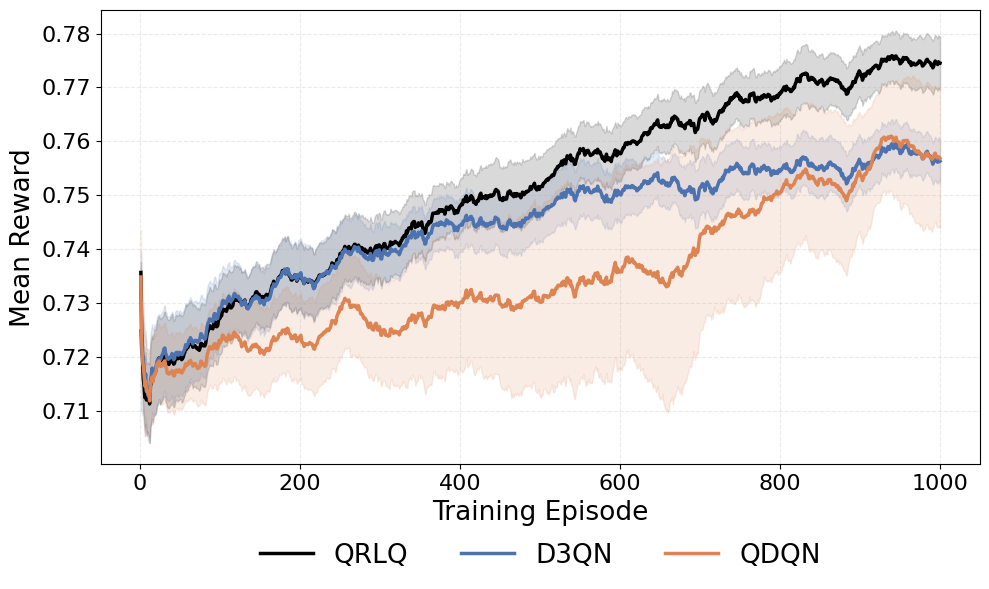}
\caption{Training curves of the mean reward.}
\label{fig:training_reward}
\end{figure}

\begin{figure*}[!t]
\centering
    \subfloat[Execution time across evaluation episodes.]{%
    \includegraphics[width=0.24\textwidth]{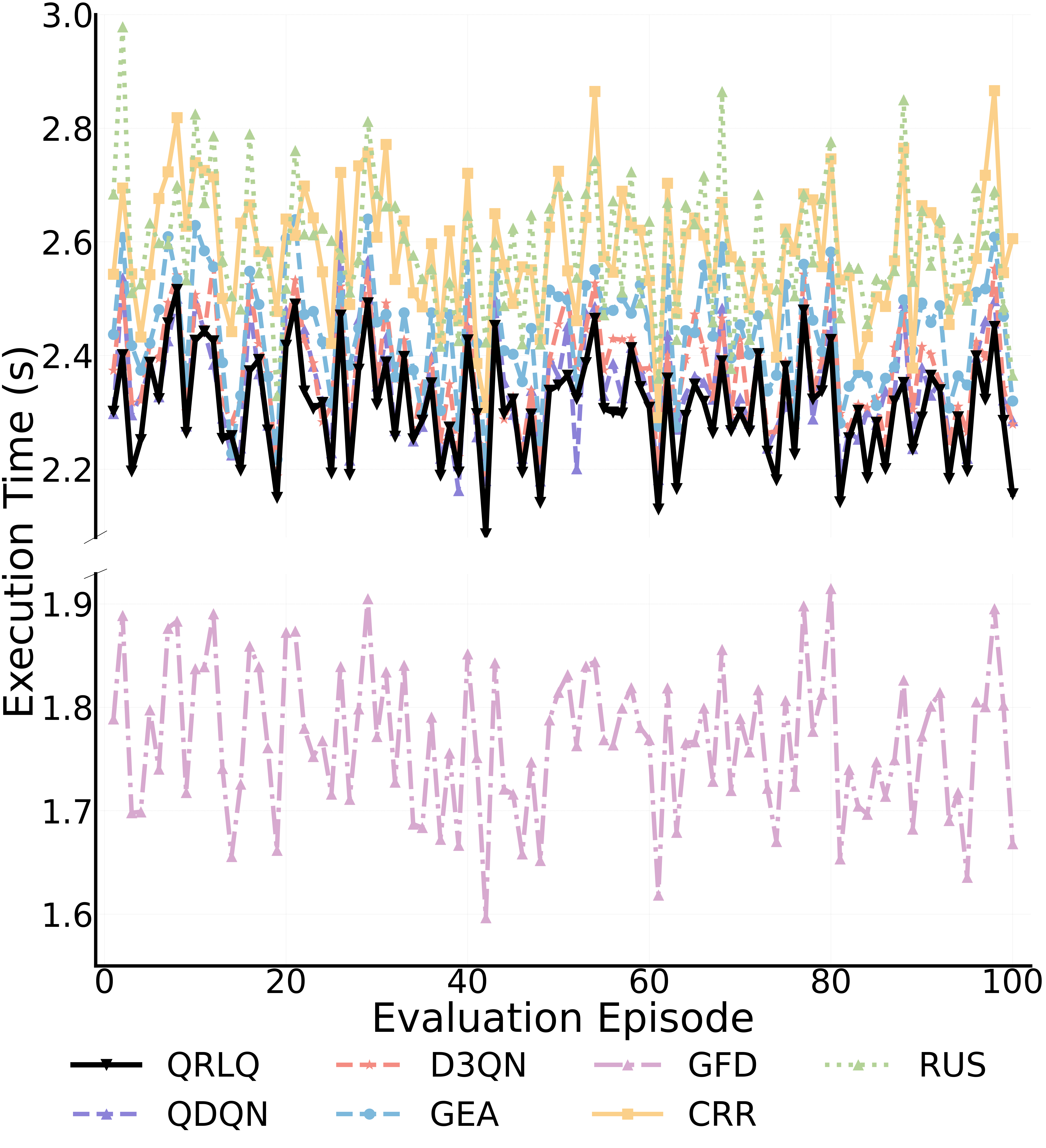}%
    \label{fig:execution_time_episodes}}
        \hspace{0.005\textwidth}
    \subfloat[Average task execution time comparison.]{%
    \includegraphics[width=0.24\textwidth]{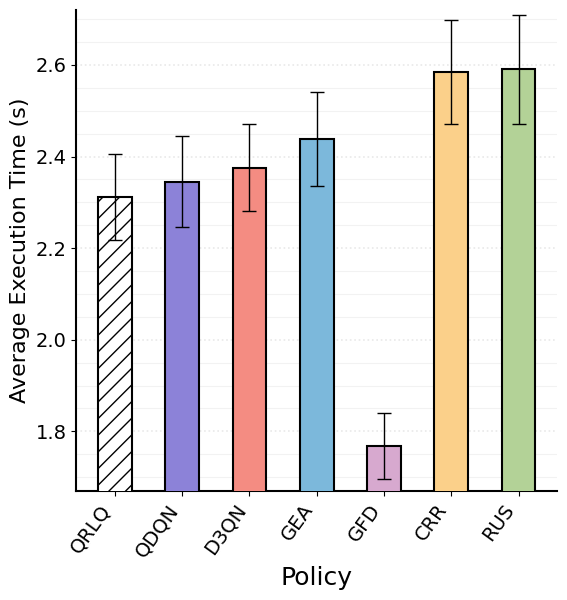}%
    \label{fig:execution_time_avg}} 
        \hspace{0.005\textwidth}
    \subfloat[Waiting time across evaluation episodes.]{%
    \includegraphics[width=0.24\textwidth]{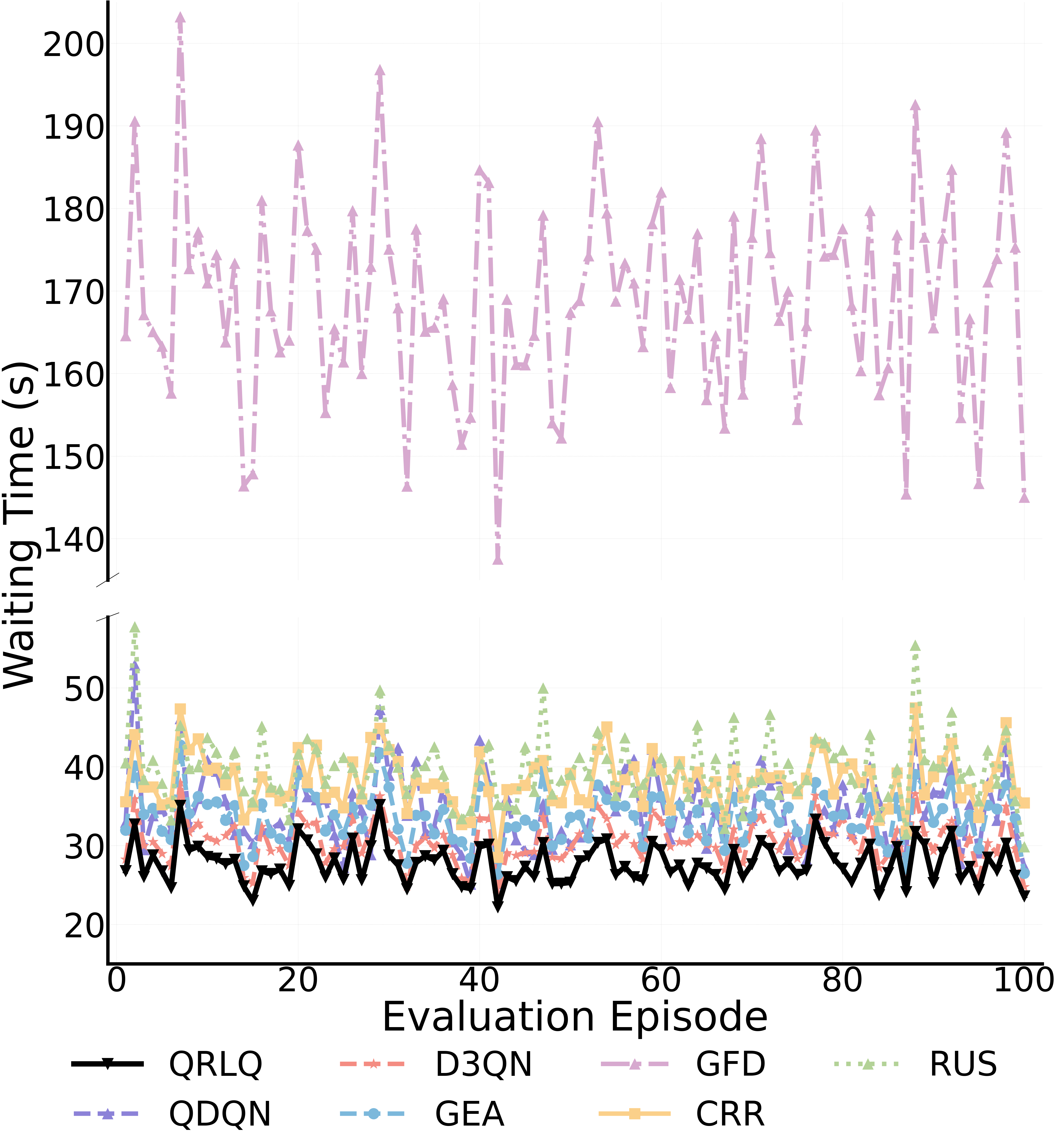}%
    \label{fig:waiting_time_episodes}}
        \hspace{0.005\textwidth}
    \subfloat[Average task waiting time comparison.]{%
    \includegraphics[width=0.24\textwidth]{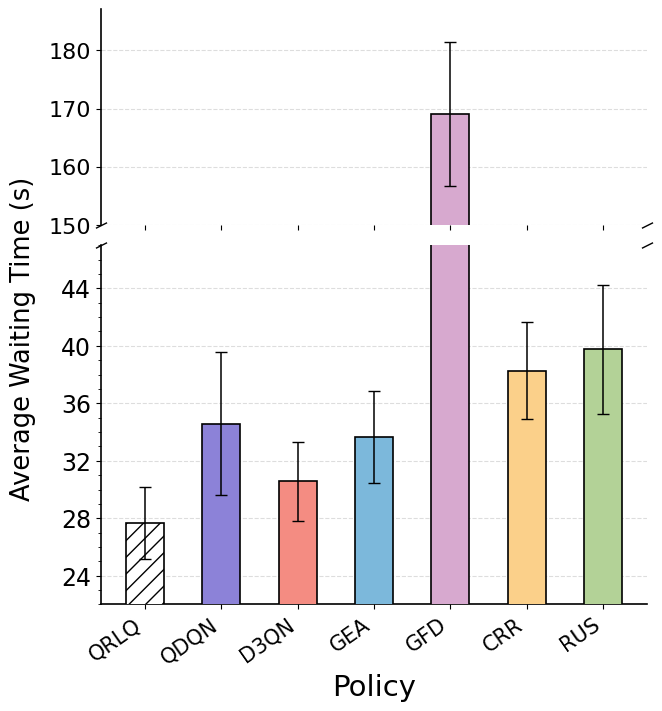}%
    \label{fig:waiting_time_avg}}
\caption{Performance comparison of execution time and waiting time across policies. Subfigures (a) and (c) show the variation across evaluation episodes, while (b) and (d) report the corresponding average values with standard deviation.}
\label{fig:performance_execution_waiting}
\end{figure*}

\section{Performance Evaluation}
\label{sec:performance_evaluation}

\subsection{Simulation Setup}
We set up a heterogeneous quantum cloud simulation environment using the QSimPy simulation tool \cite{nguyen2025qsimpy}. The quantum computing tasks are generated from the MQT Bench library \cite{quetschlich2023mqt}, including more than 20 standard algorithms with abstract circuit depths ranging from 3 to 30 layers and requiring 2 to 27 measurement qubits. The task arrival times follow a Poisson distribution to model stochastic job submissions observed in practical quantum cloud resource environments. The simulation cluster is constructed using Qiskit fake backend instances that emulate representative IBM quantum devices. The configuration includes two 27-qubit processors representing \texttt{ibm\_sydney} and \texttt{ibm\_kolkata} alongside three 127-qubit processors representing \texttt{ibm\_sherbrooke}, \texttt{ibm\_washington}, and \texttt{ibm\_brisbane}. The QRLQ framework is implemented using the TorchQuantum library \cite{wang2022quantumnas} to enable the training of a PQC on classical hardware. Simulating an idealised environment lets us isolate algorithmic feasibility from hardware-induced errors \cite{dai2025quantum, wei2024quantum}. All simulations were conducted on an Apple M1 Pro–based system with a 10-core CPU, 16-core GPU, 16-core neural engine, and 16 GB unified memory.

To evaluate the QRLQ framework, we establish comprehensive baselines that encompass classical scheduling heuristics \cite{li2024moirai, 10.1145/3799898, 10.1145/3754598.3754641} and representative classical and quantum DRL approaches, as detailed below.

\begin{itemize}
    \item \textbf{Greedy Earliest Availability (GEA)} prioritises the earliest available resources.
    \item \textbf{Greedy Fastest Duration (GFD)} prioritises hardware with the fastest average gate execution duration.
    \item \textbf{Cyclic Round Robin (CRR)} employs a cyclic approach for allocation.
    \item \textbf{Random Uniform Selection (RUS)} performs random allocation of QTasks.
    \item \textbf{Duelling Double DQN (D3QN)} serves as a standard representative for classical DRL methods.
    \item \textbf{Quantum DQN (QDQN)} serves as a direct baseline to evaluate the proposed hybrid architecture.
\end{itemize}

Table \ref{tab:hyperparameter_settings} summarises the hyperparameter settings across the evaluated policies. All three models use the same set of core reinforcement learning hyperparameters. Although the learning rate is kept constant, the QDQN model requires an additional scaling parameter to map the raw expectation values produced by the quantum circuit from the interval $[-1, 1]$ to approximate the target Q-value range required for action selection \cite{skolik2022quantum}.

\subsection{Performance Study}
\subsubsection{QRLQ Training Performance and Efficiency}

We set $\beta=0.5$ for all trained policies to assign equal importance to execution time and waiting time in the reward function. This configuration evaluates the proposed framework under a balanced objective. Fig.~\ref{fig:training_reward} shows the mean reward over 1000 training episodes, averaged across five independent random seeds. The shaded regions represent the corresponding 95\% confidence intervals. During the first 200 episodes, all three reinforcement learning methods rapidly improved their rewards, reflecting effective policy learning in the early training stage. Among them, QRLQ consistently achieved higher mean rewards while maintaining relatively narrower confidence intervals than the baseline methods. After this stage, the QDQN model continued to improve from a lower reward level and gradually approached the performance of D3QN toward the end of training, although it exhibited larger confidence intervals throughout most of the training process. By comparison, the D3QN model continued to improve at a slower rate after its initial growth and eventually plateaued at a lower reward level than QRLQ. In contrast, the QRLQ framework steadily increased its average reward throughout training and stabilised at approximately 0.775 during the final 100 episodes, with only minor fluctuations during the final stage of training. Overall, the QRLQ framework achieved the highest mean reward and exhibited narrower confidence intervals than the baseline methods during the final 100 training episodes.

\subsubsection{Execution and Waiting Time Performance}
We evaluated the QRLQ policy trained with random seed 21 on 21,000 independent QTasks over 100 evaluation episodes, similarly for the learning-based baselines. These QTasks were not used during training, allowing us to evaluate generalisation on unseen workloads. Fig.~\ref{fig:execution_time_episodes} shows the average execution time across evaluation episodes. The execution times of QRLQ remained within a relatively narrow range across the evaluated episodes. This narrow range indicates that QRLQ produced consistent execution-time performance across different evaluation episodes. QRLQ was second only to GFD, which mostly ranged from 1.6 to 1.9 seconds, and outperformed the remaining heuristic algorithms, including GEA, CRR, and RUS. Fig.~\ref{fig:execution_time_avg} shows the overall average execution time, with error bars indicating the standard deviation. The average results follow the same trend, with QRLQ outperforming all evaluated heuristic strategies except GFD. Compared with the learning-based baselines, the QRLQ framework achieved execution times comparable to those of the QDQN and D3QN models. It reduced execution time by 1.46\% compared with QDQN and by 2.70\% compared with D3QN, while using substantially fewer trainable parameters. Although GFD achieved the lowest execution time by selecting the fastest available QPU, this greedy selection can concentrate tasks on a small subset of devices, creating longer queues for subsequent tasks.

Figs.~\ref{fig:waiting_time_episodes} and \ref{fig:waiting_time_avg} present the variation in average waiting time per episode and the overall average across 100 evaluation episodes, respectively. As shown in Fig.~\ref{fig:waiting_time_episodes}, the local greedy objective of GFD led to queue congestion, resulting in high waiting times that mostly ranged from 148 to 210 seconds. Although the GEA strategy initially reduced waiting times by prioritising early availability, its limited look-ahead capability did not fully account for heterogeneous task complexities and varying shot requirements, leading to less effective global scheduling. The proposed QRLQ framework reduced these bottlenecks by considering both execution time and waiting time. As shown in Fig.~\ref{fig:waiting_time_avg}, QRLQ maintained a waiting time consistently under 35 seconds, outperforming all heuristic baselines, including GEA. Compared with the learning-based baselines, QRLQ yielded a lower average waiting time in the evaluated setting. While the D3QN and QDQN models also avoided heavy congestion, QRLQ achieved a moderate reduction in waiting times with smaller variance, suggesting improved stability of its hybrid scheduling policy. 

\begin{table}[htbp]
\renewcommand{\arraystretch}{1.3}
\centering
\caption{Average cost and delay per QTask with standard error across policies.}
\label{tab:cost_delay}
\begin{tabular}{lcc}
\toprule
\textbf{Policy} & \textbf{Average cost (\$)} & \textbf{Average delay (s)} \\
\hline
GEA    & 3.90 $\pm$ 0.03 & 36.09 $\pm$ 0.15 \\
GFD    & 2.83 $\pm$ 0.03 & 170.81 $\pm$ 0.69 \\
CRR    & 4.14 $\pm$ 0.04 & 40.86 $\pm$ 0.22 \\
RUS    & 4.14 $\pm$ 0.04 & 42.34 $\pm$ 0.23 \\
QDQN & 3.75 $\pm$ 0.03 & 36.92 $\pm$ 0.30 \\
D3QN & 3.80 $\pm$ 0.04 & 32.94 $\pm$ 0.14 \\
\textbf{QRLQ}   & \textbf{3.70 $\pm$ 0.03} & \textbf{29.99 $\pm$ 0.14} \\
\toprule
\end{tabular}
\end{table}

\subsubsection{Cost–Delay Tradeoff Analysis}
Based on the preceding analysis of execution time and waiting time, this section examines their impact on cost and delay. To calculate cost, we applied the IBM Quantum on-demand price of \$96 per minute, which equates to \$1.60 per second of actual execution time on physical quantum hardware. Table~\ref{tab:cost_delay} reports the average cost and delay per QTask. Consistent with the execution time results, the GFD strategy recorded the lowest cost at \$2.83. However, Table \ref{tab:cost_delay} also shows a limitation of this strategy in terms of delay. Because of the long waiting times discussed previously, GFD had the highest average delay of 170.81 seconds, showing that a low execution cost does not necessarily lead to a low completion time. In contrast, QRLQ achieved the lowest mean delay of 29.99 seconds while incurring the second-lowest mean cost of \$3.70 per QTask among the evaluated policies. Its delay was lower than that of all heuristic approaches, including the GEA strategy that explicitly prioritises the earliest available resources. In addition, QRLQ achieved both lower cost and shorter delay than the learning-based baselines, including QDQN at \$3.75 and 36.92 seconds and D3QN at \$3.80 and 32.94 seconds.

\begin{figure}[htbp]
\centering
\hspace{-0.47cm}
\includegraphics[width=0.72\linewidth]{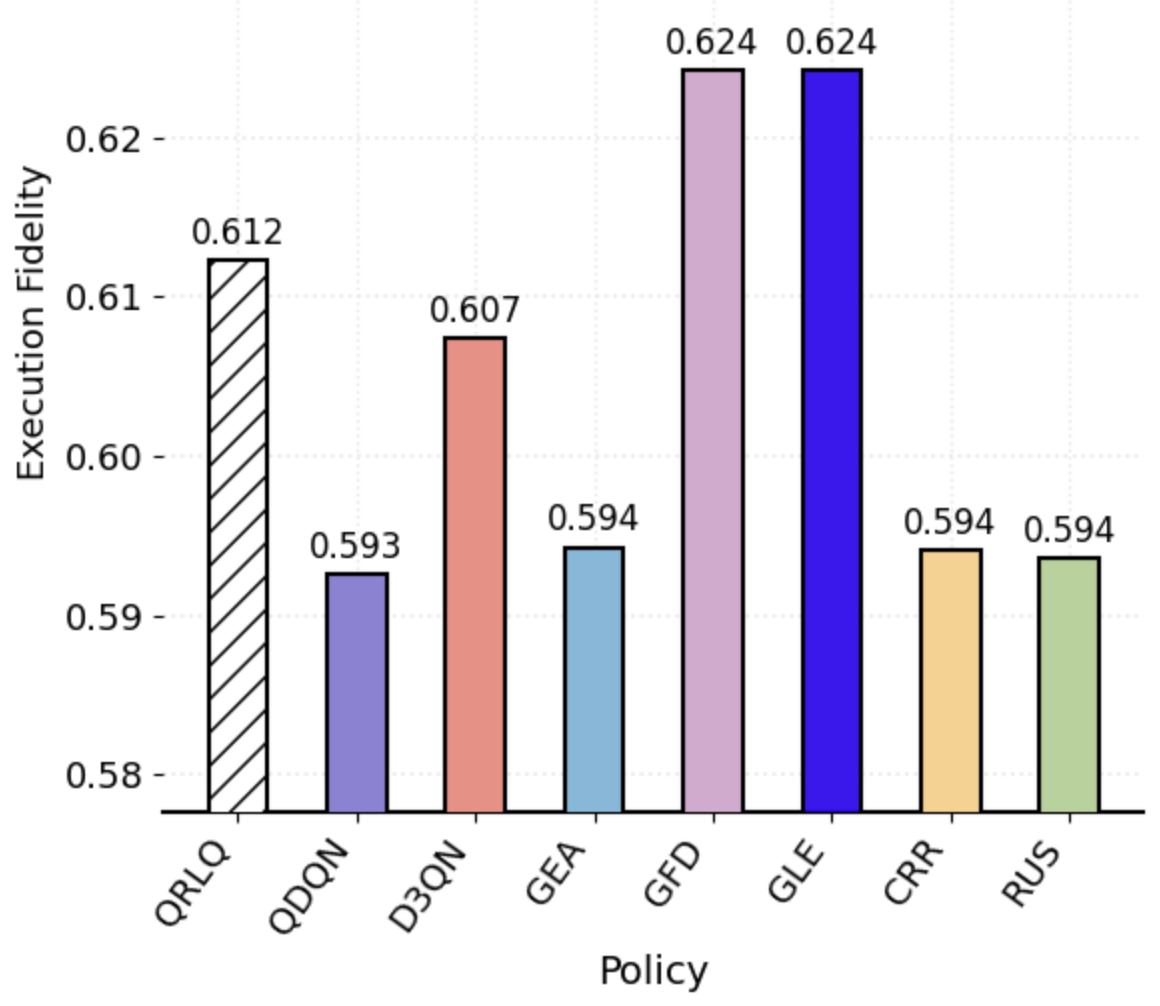}
\caption{Average execution fidelity across evaluated policies.}
\label{fig:execution_fidelity_avg}
\end{figure}

\subsubsection{Execution Fidelity Performance Analysis}
Execution fidelity remains an important factor in the NISQ era \cite{10.1145/3799898}. To further evaluate the proposed strategy, an additional baseline denoted as Greedy Lowest Error (GLE) was added to select the QNodes with the lowest average gate error rate within the system. Fig.~\ref{fig:execution_fidelity_avg} shows that both the speed-prioritised GFD strategy and the error-prioritised GLE strategy achieved the highest execution fidelity, with an identical value of 0.624. This result suggests that in the evaluated quantum cloud setting, scheduling quantum computing tasks on higher-quality QNodes can achieve high execution fidelity while also maintaining a short execution time. In comparison, QRLQ achieved a mean execution fidelity of 0.612, which was approximately 1.9\% below the highest observed value of 0.624 and remained consistently higher than the traditional D3QN and baseline QDQN models, which recorded fidelities of 0.607 and 0.593, respectively. Furthermore, the proposed approach outperformed heuristic strategies that do not explicitly consider hardware quality, such as GEA, CRR, and RUS. Across the evaluated policies, QRLQ achieved the lowest mean delay, a lower mean cost than GEA, CRR, and RUS, and a mean execution fidelity close to the highest observed value.

\subsubsection{Computational Complexity and Efficiency Analysis}
\label{sec:computational_complexity}
We evaluated the computational efficiency of the proposed models by analysing the time complexity with respect to several key parameters. Specifically, $n$ denotes the number of qubits, and $l$ represents the number of quantum layers. The parameters $a$ and $b$ indicate the number of variational gates and data embedding gates operating on each qubit within a single layer \cite{dai2025quantum}. Finally, $P_c$ represents the total number of trainable parameters of the classical neural network in the hybrid architecture. In this work, we configured $n$, $a$, $b$, $l$, and $P_c$ to be 8, 2, 3, 1, and 198, respectively.

When analysing the time complexity of execution on quantum hardware, the computational cost is determined by both quantum and classical components. Within each quantum layer, data embedding gates and variational gates operate independently on each qubit, whereas entanglement gates require sequential operation. This structure results in a complexity of $O(a+b+n)$ per quantum layer and $O(l \times (a+b+n))$ across $l$ quantum layers. The output from the quantum circuit then passes through a classical neural network, so the overall time complexity is approximated as $O(l \times (a+b+n) + P_c)$. When simulating the algorithm on a classical computer, the computational cost increases exponentially because every quantum operation updates the state vector in a Hilbert space of size $2^n$. The total number of quantum gates in the circuit is calculated as $l \times \left((a+b)\times n + n\right)$. Therefore, the simulation time complexity for the quantum component alone is expressed as $O(l \times \left((a+b)\times n + n\right) \times 2^n)$. The classical computation associated with $P_c$ is then added, yielding an overall simulation time complexity of $O(l \times \left((a+b)\times n + n\right) \times 2^n + P_c)$.

Regarding memory requirements during execution on quantum hardware, the model requires memory allocation for the trainable quantum rotation angles, which yields a space complexity of $O(l \times a \times n)$. When combining this requirement with the trainable weights of the classical network, the overall space complexity is given by $O(l \times a \times n + P_c)$. In classical simulation, the storage space required on platforms such as TorchQuantum depends primarily on the need to maintain a complex state vector of size $2^n$ alongside the parameter count $P_c$ of the hybrid architecture. Consequently, the overall classical simulation space complexity is $O(2^n + P_c)$.

Based on the complexity analysis, the proposed QRLQ architecture requires 16 trainable quantum parameters. When combined with the 198 parameters of the classical network, the total parameter count reaches 214. Compared to the classical D3QN model comprising 758 parameters, the proposed hybrid framework uses fewer trainable parameters than the classical D3QN baseline. Compared with the QDQN baseline using a 10-layer, 8-qubit circuit, QRLQ employs a shallower quantum circuit, resulting in fewer quantum operations than the QDQN baseline. Finally, regarding classical simulation overhead, although maintaining the $2^n$ state vector typically requires exponential resources, our implementation uses only 8 qubits, yielding a state vector with 256 amplitudes. With the number of qubits fixed, the simulation cost depends mainly on $l$. Therefore, compared with the QDQN baseline, the shallower quantum circuit in QRLQ reduces the number of quantum operations to be simulated, helping balance quantum feature extraction and training time on current quantum simulation platforms. In our simulations, QRLQ required approximately 10.5 hours of training on both CPU and GPU, QDQN required approximately 11.25 hours on both CPU and GPU, and D3QN required approximately 10.5 hours on the GPU.

\begin{figure}[htbp]
\centering
\includegraphics[width=0.48156\linewidth]{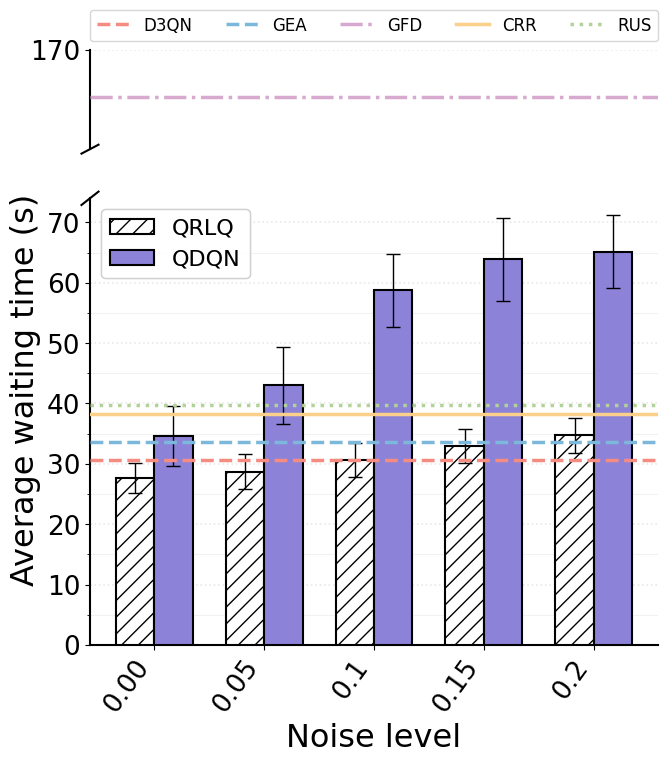}
\hfill
\includegraphics[width=0.5\linewidth]{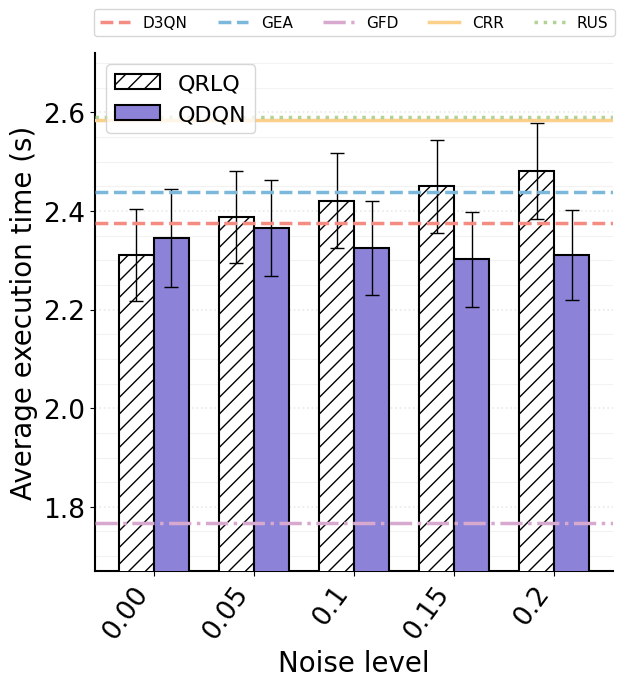}
\caption{Comparison of waiting time and execution time under different noise levels, with horizontal lines indicating the mean values of the classical baselines.}
\label{fig:noise_comparison}
\end{figure}

\subsubsection{Impact of Noise Model and Discussion}
To evaluate the scheduling performance of the trained policies under simulated noise conditions, we introduced parameter perturbations during the evaluation phase by applying random Gaussian perturbations to the variational parameters of the PQCs \cite{Andrea2022quantum}. Specifically, each trained rotation angle $\theta$ was perturbed such that $\theta_{\mathrm{noisy}} = \theta + \delta\theta$, where $\delta\theta$ was sampled from a Gaussian distribution $\mathcal{N}(0,\sigma^2)$ with $\sigma$ denoting the perturbation level. 

Fig.~\ref{fig:noise_comparison} shows that the scheduling performance of both models changed as the perturbation level increased. QRLQ exhibited relatively small changes in waiting time across the evaluated perturbation levels, whereas QDQN showed a substantially larger variation in waiting time. The execution time of both models remained within a relatively narrow range, although QRLQ exhibited a larger variation than QDQN. In terms of waiting time, QRLQ remained below most of the classical baseline means across the evaluated perturbation levels, whereas QDQN exceeded some of the classical baseline means as the perturbation level increased. For execution time, the results of both quantum models remained within the range of the classical baseline means across the evaluated perturbation levels, although QRLQ increased gradually with the perturbation level. QDQN uses a larger PQC with 160 variational gates compared with 16 in QRLQ, which may contribute to the greater sensitivity in waiting time observed under parameter perturbations. The results suggest that QRLQ maintained more consistent waiting-time performance than QDQN as the perturbation level increased, and the shallower PQC used by QRLQ may be less sensitive to the evaluated parameter perturbations.

The current simulations use five QNodes with an 8-qubit shallow-depth PQC. For larger action spaces, the architecture can be extended by increasing the number of qubits, while the existing data re-uploading mechanism allows additional input features to be encoded when needed. The classical duelling heads can also be expanded to accommodate additional scheduling actions. The appropriate quantum circuit size for larger action spaces remains to be evaluated.

\section{Conclusion and Future Work}
\label{sec:conclusion}
We have developed QRLQ, a cost-delay-aware quantum cloud scheduling framework for heterogeneous QaaS environments operating under a uniform time-based pricing model. By integrating PQCs into a D3QN, QRLQ provides a parameter-efficient reinforcement learning approach to jointly optimise execution cost and delay. Simulation results demonstrated that QRLQ achieved a 5--11\% lower mean cost relative to availability-based and rotation-based heuristics and reduced mean delay by 17\% and 82\% relative to the strongest and weakest heuristic baselines, respectively, while maintaining execution fidelity within 2\% of a fidelity-greedy policy. Compared with the classical DRL baseline, QRLQ achieved comparable scheduling performance using approximately 72\% fewer trainable parameters, suggesting the feasibility of parameter-efficient quantum-classical models for quantum cloud resource orchestration. Despite these encouraging results, several directions remain for future research. First, the current framework assumes a single quantum task per QPU, and extending QRLQ to support quantum multiprogramming would enable concurrent workload execution. Second, future work will investigate the scalability of QRLQ under larger action spaces by studying appropriate qubit counts and circuit depths. Third, integrating circuit cutting and distributed quantum execution could improve scalability beyond individual device limitations. Finally, extending the scheduling framework to iterative hybrid quantum-classical workloads, such as VQE and QAOA, and validating the proposed approach on physical quantum hardware clusters would further demonstrate its practicality under realistic NISQ conditions.

\ifCLASSOPTIONcaptionsoff
  \newpage
\fi

\bibliographystyle{IEEEtran}
\bibliography{IEEEabrv,references}
\end{document}